\documentclass{article}
\usepackage[preprint]{corl_2022}
\usepackage{multicol}
\usepackage{wrapfig}
\usepackage{times}
\usepackage{amsmath}
\usepackage{amssymb}
\usepackage{siunitx}
\usepackage{gensymb}
\usepackage{booktabs}
\usepackage{multirow}

\usepackage{todonotes}
\usepackage{subcaption}
\usepackage{xcolor}
\usepackage{adjustbox}
\usepackage{pdflscape}
\usepackage{hyperref}
\hypersetup{
	colorlinks=true,
	linkcolor=blue,
	urlcolor=blue,
}
\usepackage[subtle]{savetrees}
\usepackage{enumitem}
\setlist[itemize]{leftmargin=2em}
\setlist[enumerate]{leftmargin=2em}

\definecolor{mygreen}{rgb}{0, 0.5, 0}
\definecolor{mygrey}{rgb}{0.6, 0.6, 0.6}

\newcommand{\mypara}[1]{\vspace{-1mm}\par\vspace*{0.5mm}\noindent\textbf{{#1}}}
\newcommand{\mysection}[1]{\vspace{-3mm}\section{#1}\vspace{-3mm}}
\newcommand{\mysubsection}[1]{\vspace{-3mm}\subsection{#1}\vspace{-3mm}}

\newcommand{\ie}{\textit{i}.\textit{e}., }
\newcommand{\eg}{\textit{e}.\textit{g}. }
\newcommand{\img}{\mathcal{I}}
\newcommand{\textlabel}{\mathcal{T}}
\newcommand{\relevancy}{\mathcal{R}}
\newcommand{\occupency}{\mathcal{O}}
\newcommand{\relevancypcd}{\relevancy^\mathrm{proj}}
\newcommand{\relevancyvol}{\relevancy^\mathrm{vox}}

\newcommand{\threeDNet}{f_\mathrm{encode}}
\newcommand{\ImplicitDecoder}{f_\mathrm{decode}}
\newcommand{\spatialEmbeddingfunction}{f_\mathrm{spatial}}

\newcommand{\threeDFeatureVolume}{Z}

\newcommand{\spatialrelation}{\mathcal{S}}

\def\ours{Semantic Abstraction}
\def\firstmodule{semantic-aware wrapper}
\def\secondmodule{semantic-abstracted 3D module}

\title{
   Semantic Abstraction:
   Open-World 3D Scene Understanding from 2D Vision-Language Models
}

\author{
   Huy Ha \qquad \qquad Shuran Song\\
   Columbia University \\
   \href{semantic-abstraction.cs.columbia.edu}{semantic-abstraction.cs.columbia.edu}
}

\begin{document}

\maketitle
\vspace{-10mm}
\begin{abstract}
	We study open-world 3D scene understanding, a family of tasks that require agents to reason about their 3D environment with an open-set vocabulary and out-of-domain visual inputs --  a critical skill for robots to operate in the unstructured 3D world. Towards this end, we propose Semantic Abstraction (SemAbs), a framework that equips 2D Vision-Language Models (VLMs) with new 3D spatial capabilities, while maintaining their zero-shot robustness. We achieve this abstraction using relevancy maps extracted from CLIP, and learn 3D spatial and geometric reasoning skills on top of those abstractions in a semantic-agnostic manner. We demonstrate the usefulness of SemAbs on two open-world 3D scene understanding tasks: 1) completing partially observed objects and 2) localizing hidden objects from language descriptions. SemAbs can generalize to novel vocabulary for object attributes and nouns, materials/lighting, classes, and domains (i.e., real-world scans) from training on limited 3D synthetic data.
\end{abstract}
\vspace{-4mm}


\begin{figure}[h]
	\vspace{-2mm}
	\centering
	\includegraphics[width=\linewidth]{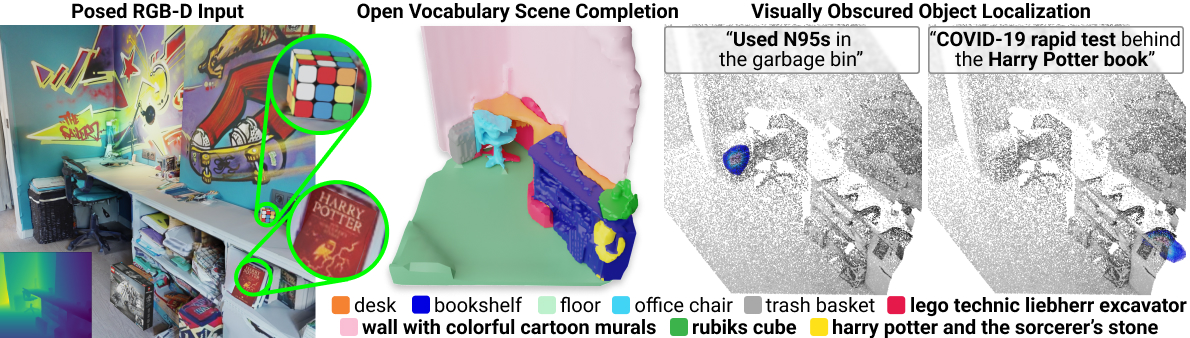}
	\caption{ \footnotesize
		\textbf{Open-World 3D Scene Understanding}.
		Our approach, Semantic Abstraction, unlocks 2D VLM's capabilities to 3D scene understanding.
		Trained with a small simulated data, our model generalizes to unseen classes in a novel domain (\ie real world scans), from small objects like ``rubiks cube'', to long-tail concepts like ``harry potter'', to hidden objects like the ``used N95s in the garbage bin''.
	}
	\label{fig:teaser}
	\vspace{-3mm}
\end{figure}
\mysection{Introduction}
To assist people in their daily life, robots must recognize a large,
context-dependent, and dynamic set of semantic categories from their visual
sensors. However, if their 3D training environment contains only \emph{common}
categories like desks, office chairs, and bookshelves, how could they represent
and localize \emph{long-tail} objects, such as the ``used N95s in the garbage
bin'' (Fig~\ref{fig:teaser})?

\begin{figure}[t]
	\centering
	\includegraphics[width=\linewidth]{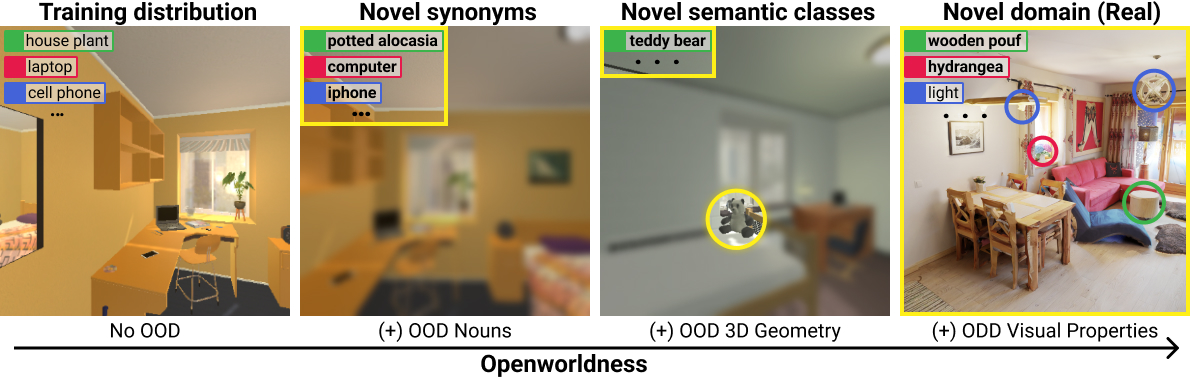}
	\vspace{-5mm}
	\caption{
		\textbf{Open-world generalization requirements} can build on top of each other, forming tiers of open-worldness (distinct property of each tier outlined yellow).
	} \vspace{-6mm}
	\label{fig:openworld}
\end{figure}

This is an example of an open-world 3D scene understanding task
(Fig.~\ref{fig:openworld}), a family of 3D vision-language tasks which
encompasses open-set classification (evaluation on novel classes) with extra
generalization requirements to novel vocabulary (\ie object attributes,
synonyms of object nouns), visual properties (\eg lighting, textures), and
domains (\eg sim v.s. real). The core challenge in such tasks is the limited
data: all existing 3D
datasets~\cite{ai2thor,song2015sun,chang2015shapenet,ramakrishnan2021hm3d,szot2021habitat,chen2020scanrefer}
are limited in diversity and scale compared to their internet-scale 2D
counterparts~\cite{radford2021learning,li2021align,alayrac2022flamingo}, so
training on them does not prepare robots for
the open 3D world.

On the other hand, large-scale 2D vision language models
(VLM)~\cite{radford2021learning,li2021align,pham2021combined,alayrac2022flamingo,jia2021scaling}
have demonstrated impressive performance on open-set image classification.
Through exposure to many image-caption pairs from the internet, they have
acquired an unprecedented level of robustness to visual distributional shifts
and a large repertoire of visual-semantic concepts and vocabulary.
However, fine-tuning these pretrained models reduces their open-set
classification
robustness~\cite{radford2021learning,pham2021combined,wortsman2021robust,andreassen2021evolution}.
Compared to internet-scale image-caption data, the finetuning datasets,
including the existing 3D
datasets~\cite{ai2thor,song2015sun,chang2015shapenet,ramakrishnan2021hm3d,szot2021habitat,chen2020scanrefer},
show a significant reduction in scale and diversity, which causes finetuned
models to become task-specialized and lose their generality. Hence, we
investigate the following question:

\begin{quote}
	\vspace{-3mm}
	\centering
	\it{
		How can we equip 2D VLMs with new 3D capabilities,
		\\
		while maintaining their zero-shot robustness?}
	\vspace{-3mm}
\end{quote}


We propose \textbf{\ours~(SemAbs)}, a framework for tackling visual-semantic
reasoning in open-world 3D scene understanding tasks using 2D VLMs. We
hypothesize that, while open-world visual-semantic reasoning requires exposure
to internet-scale datasets, 3D spatial and geometric reasoning is tractable
even with a limited synthetic dataset, and could generalize better if learned
in a semantic-agnostic manner.
For instance, instead of learning the concept of ``behind the Harry Potter
book'', the 3D localization model only needs to learn the concept of ``behind
\emph{that object}''. 

To achieve this abstraction, we leverage the relevancy maps extracted from 2D
VLMs and used them as the ``abstracted object'' representation that is agnostic
to their semantic labels. The SemAbs module factorizes into two submodules: 1)
A \firstmodule~that takes an input RGB-D image and object category label and
outputs the relevancy map of a pre-trained 2D VLM model (i.e., CLIP
\cite{radford2021learning}) for that label, and 2) a {\secondmodule~} that uses
the relevancy map to predict 3D occupancy. This 3D occupancy can either
represent the 3D geometry of a partially observed object or the possible 3D
locations of a hidden object (\eg mask in the trash) conditioned on the
language input.

While we only train the 3D network on a limited synthetic 3D dataset, it
generalizes to any novel semantic labels, vocabulary, visual properties, and
domains that the 2D VLM can generalize to. As a result, the SemAbs module
inherits the VLM's visual robustness and open-set classification abilities
while adding the spatial reasoning abilities it
lacks~\cite{subramanian2022reclip,gadre2022clip,ramesh2022hierarchical}. In
summary, our contributions are three-fold\footnote{All code, data, and models
	is publicly available at
	\href{https://github.com/columbia-ai-robotics/semantic-abstraction}{https://github.com/columbia-ai-robotics/semantic-abstraction}.}:
\begin{itemize}\setlength\itemsep{-0.0em}
	\vspace{-2mm}
	\item \textbf{\ours}, a framework for augmenting 2D VLMs with 3D reasoning capabilities for open-world 3D scene understanding tasks.
	      By abstracting semantics away using relevancy maps,
	      our SemAbs module generalizes to novel semantic labels, vocabulary, visual properties, and domains (\ie sim2real) despite being trained on only a limited synthetic 3D dataset.

	\item \textbf{Efficient Multiscale Relevancy Extraction.} To support \ours, we propose a multi-scale relevancy extractor for vision transformers~\cite{dosovitskiy2020image} (ViTs), which robustly detects small, long-tail objects and achieves over $\times\mathbf{60}$ speed up from prior work~\cite{chefer2021generic}.

	\item Two novel \textbf{Open-world 3D Scene Understanding} tasks (open-vocabulary
	      semantic scene completion and visually obscured object localization), a data
	      generation pipeline with AI2-THOR~\cite{ai2thor} simulator for the tasks, and a
	      systematic open-world evaluation procedure. We believe these two tasks are
	      important primitives for bridging existing robotics tasks to the open-world.
	      \vspace{-2mm}
\end{itemize}
\mysection{Related Works}

\mypara{2D Visual Language Models.}
The recent advancements in scaling up contrasting learning have enabled the training of large 2D vision language models (VLM).
Through their exposure to millions of internet image-caption pairs, these VLMs \cite{radford2021learning,pham2021combined,alayrac2022flamingo,jia2021scaling} acquire a remarkable level of zero-shot robustness -- they can recognize a diverse set of long-tail semantic concepts robustly under distributional shifts.
However, the learned image-level representations are not directly applicable to spatial-reasoning tasks which require pixel-level information.
To address this issue, approaches were proposed to extract dense features~\cite{zhou2021denseclip,rao2022denseclip,li2022adapting} or relevancy maps \cite{chefer2021generic,selvaraju2017grad}.
These advancements complement our contribution, which leverages such techniques for interfacing 2D VLMs to open-world 3D scene understanding tasks.

\mypara{Transferring 2D VLMs to 2D Applications.}
Encouraged by their impressive cabilities, many downstream tasks build around these pretrained VLMs~\cite{mokady2021clipcap,song2022clip,gao2021clip2tv,subramanian2022reclip,wang2021actionclip,zabari2021inthewild,shridhar2021cliport}, often finetuning them (either end-to-end or train learnable weights on top of the pretrained encoder) on a small task-specific dataset.
However, many recent studies show that finetuning these VLMs significantly weakens their zero-shot robustness~\cite{radford2021learning,wortsman2021robust,pham2021combined,andreassen2021evolution,zhou2021denseclip}, which motivates a new paradigm for using pretrained models.
By combining a \textit{fixed} pretrained VLM  with prompting~\cite{zeng2022socratic} or attention extraction~\cite{zhou2021denseclip,rao2022denseclip,li2022adapting,gadre2022clip}, the new generation of zero-shot transfer techniques  can inherit the VLM's robustness in their downstream tasks without overfitting to any visual domain.
However, all existing zero-shot transfer techniques have been confined to the realms of 2D vision.

\mypara{Closed-World 3D Scene Understanding.}
Understanding the semantics and 3D structure of an unstructured environment is a fundamental capability for robots.
However, prior works in 3D object detection~\cite{SlidingShapes,DSS,hou20193dsis}, semantic scene completion~\cite{song2017semantic,dai2018scancomplete,avetisyan2019scan2cad}, and language-informed object localization~\cite{chen2020scanrefer, roh2022languagerefer} are typically concerned with limited visual diversity and object categories, limitations imposed by their 3D training data.
Despite great efforts from the community~\cite{ai2thor,song2015sun,chang2015shapenet,ramakrishnan2021hm3d,szot2021habitat,chen2020scanrefer}, existing 3D datasets' scale, diversity, and coverage pale in comparison to internet-scale image-text pairs~\cite{radford2021learning,li2021align,alayrac2022flamingo}.
However, our key insight is that visual-semantic reasoning can be learned separately from 3D reasoning.
That is, with the right abstraction, the complex visual-semantic reasoning can be offloaded to 2D VLMs, while the 3D model specializes in semantic-agnostic spatial and geometry reasoning.

\mysection{Method: Semantic Abstraction}
\label{sec:method}

Evidence from linguistics and cognitive science suggests that semantic concepts are much more diverse than spatial and geometric concepts.
While the average adult knows roughly 10,000 nouns for object categories, there are only ~90 words in English for describing spatial relations~\cite{barbara1993whatwhere}.
In fact, there are so few spatial prepositions that they are typically considered a closed-class vocabulary~\cite{hayward1995spatial}.
Similarly for 3D geometry, the idea that our 3D world can be decomposed into sets of primitives~\cite{tulsiani2017learning} has been applied in many works, such as generalized cylinders~\cite{binford1971visual} and block worlds~\cite{gupta2010blocks}.
In contrast, the space of nouns for semantic concepts is large and often cannot be further decomposed into common primitives.
This suggests that while we may need a large and diverse dataset for semantic and visual concept learning, a relatively small 3D dataset could cover spatial and geometric reasoning.
This observation motivates our approach, Semantic Abstraction.

\mysubsection{Abstraction via Relevancy}
\label{sec:method:semabs}

\begin{figure}[t]
	\centering
	\includegraphics[width=\linewidth]{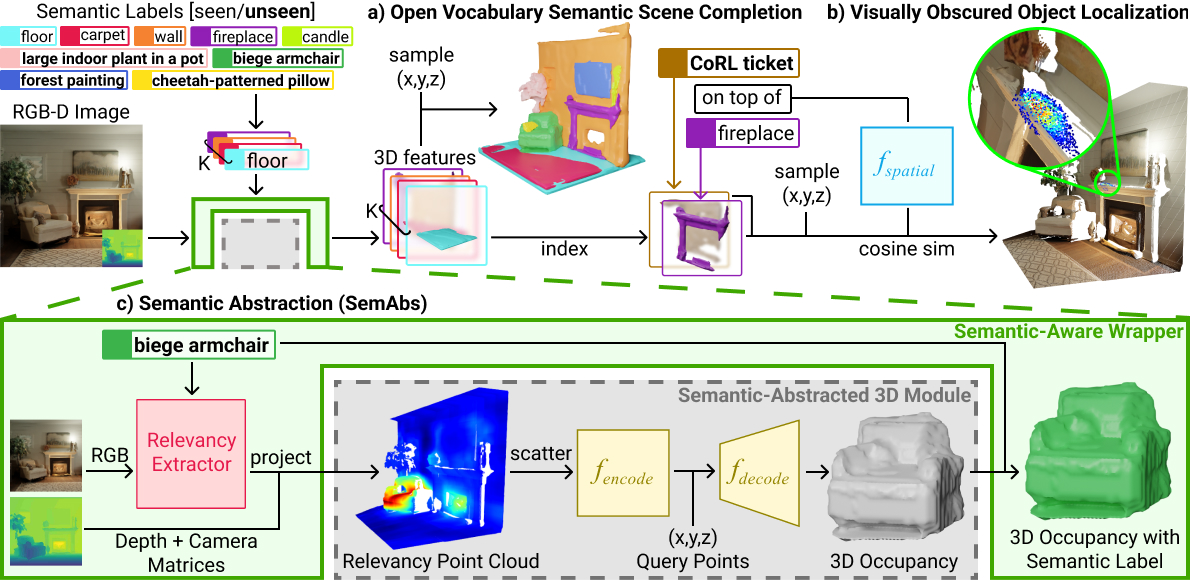}
	\caption{
		\footnotesize
		\textbf{Semantic Abstraction Overview.}
		Our framework can be applied to open-world 3D scene understanding tasks (a-b) using the SemAbs module (c).
		It consists of a \textbf{\firstmodule} (green background) that abstracts the input image and semantic label into a relevancy map,
		and a \textbf{\secondmodule} (grey background) that completes the projected relevancy map into a 3D occupancy.
		This abstraction allows our approach to generalize to long-tail semantic labels unseen (bolded) during 3D training, such as the ``CoRL ticket on top of the fireplace''.}
	\label{fig:method}
	\vspace{-8mm}
\end{figure}

The Semantic Abstraction (SemAbs) module (Fig~\ref{fig:method}c) takes as input
an RGB-D image $\img \in \mathbb{R}^{H\times W}$ and an object class text label
$\textlabel$ (\eg ``biege armchair'') and outputs the 3D occupancy $\occupency$
for objects of class~$\textlabel$.
It factorizes into two submodules:

\mypara{The \firstmodule} (Fig~\ref{fig:method}c, green background) abstracts $\img$ and $\textlabel$ into a relevancy map $\relevancy \in \mathbb{R}^{H\times W}$, where each pixel's value denotes that pixel's contribution to the VLM's classification score for $\textlabel$.
Introduced for model explainability \cite{selvaraju2017grad,chefer2021generic}, relevancy maps can be treated as a coarse localization of the text label.
Using the depth image and camera matrices, $\relevancy$ is projected into a 3D point cloud, $\relevancypcd =\{r_i\}^{H\times W}_{i=1}$ where $r_i \in \mathbb{R}^4$ (a 3D location with a scalar relevancy value).
Only $\relevancypcd$ , but neither the text $\textlabel$ nor the image $\img$, is passed to the second submodule.

\mypara{The \secondmodule} (Fig~\ref{fig:method}c, grey background) treats the relevancy point cloud as the localization of a partially observed object and completes it into that object's 3D occupancy.
To do this, we first scatter $\relevancypcd$ into a 3D voxel grid $\relevancyvol \in \mathbb{R}^{D \times 128 \times 128 \times 128} $.
Then, we encode  $\relevancyvol$ as a 3D feature volume: $\threeDNet(\relevancyvol) \mapsto \threeDFeatureVolume \in \mathbb{R}^{D \times 128 \times 128 \times 128}$, where $\threeDNet$ is a 3D UNet~\cite{abdulkadir2016unet3d} with 6 levels.
$\threeDFeatureVolume$ can be sampled using trilinear interpolation to produce local features $\phi^\threeDFeatureVolume_q \in \mathbb{R}^D$ for any 3D query point $q \in \mathbb{R}^3$.
Finally, decoding $\phi^\threeDFeatureVolume_q$ with a learned MLP $\ImplicitDecoder$ gives us an occupancy probability for each point $q$, $\ImplicitDecoder\big(\phi^\threeDFeatureVolume_q\big) \mapsto o(q) \in [0,1]$.
In this submodule, only $\threeDNet$ and $\ImplicitDecoder$ (Fig.~\ref{fig:method}c, yellow boxes) are trained with the 3D dataset.

Although the \secondmodule~only observes relevancy abstractions of the semantic label $\textlabel$ but not $\textlabel$ itself, the \secondmodule's output can be interpretted as the 3D occupancy for $\textlabel$.
This means it generalizes to any semantic label that can be recognized by the 2D VLMs' relevancy maps even if it was trained on a limited 3D dataset.
In our implementation, we use CLIP~\cite{radford2021learning} as our VLM.
However, our framework is \emph{VLM-agnostic}, as long as relevancy maps can be generated.
It is a interesting future direction to investigate how different VLMs~\cite{radford2021learning,li2021align,pham2021combined,alayrac2022flamingo,jia2021scaling}, their training procedure~\cite{chefer2022optimizing}, and different relevancy approaches~\cite{chefer2021generic,chefer2021transformer,selvaraju2017grad,liu2022rethinking} affect the performance of different downstream 3D scene understanding tasks.
In the next sections, we explain how to ensure the visual-semantic concepts are
reliably recognized by the relevancy maps (Sec.~\ref{sec:method:relevancy}),
and how to apply SemAbs to 3D scene understanding tasks
(Sec.~\ref{sec:method:ovssc}, Sec.~\ref{sec:method:vool}).

\mysubsection{A Multi-Scale Relevancy Extractor}
\label{sec:method:relevancy}

\begin{wrapfigure}{r}{0.47\textwidth}
	\centering
	\includegraphics[width=\linewidth]{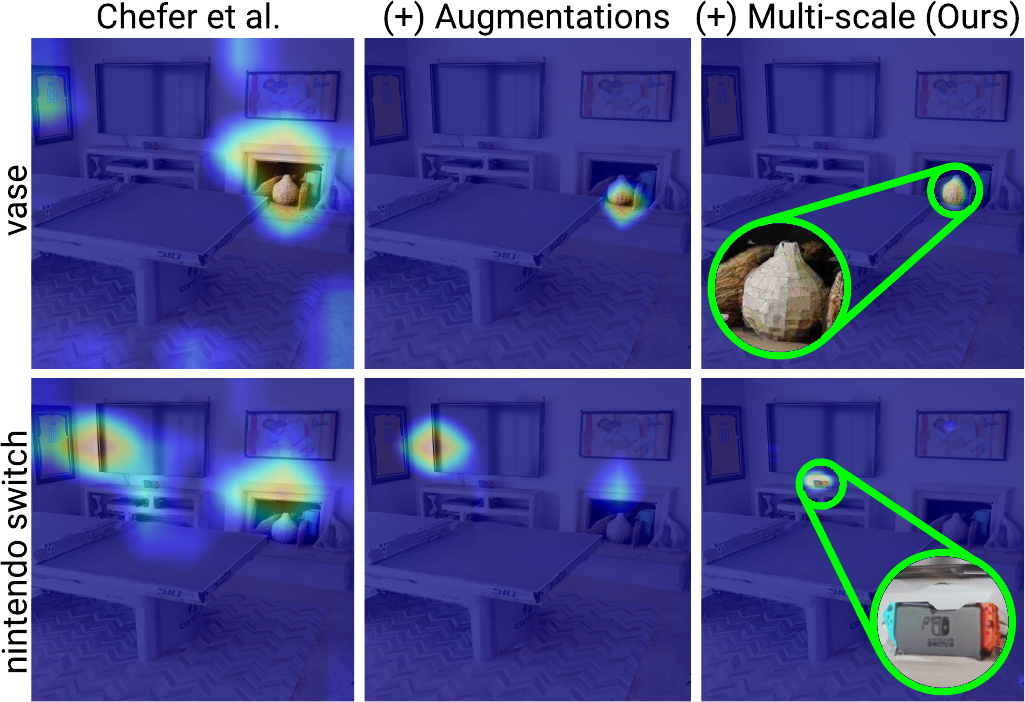}
	\caption{ \footnotesize
		Our relevancy extractor robustly detects even small, long-tail objects, like
		the ``nintendo switch''.} \vspace{-8mm} \label{fig:relevancy_improvements}
\end{wrapfigure}

We choose to use ViT-based CLIP~\cite{radford2021learning} due to their
superior performance over the ResNet variants. However, existing ViT relevancy
techniques~\cite{chefer2021generic} often produce noisy, low-resolution maps
that highlight irrelevant regions or miss small objects
\cite{shen2021much,zabari2021inthewild}.

To reduce noise from irrelevant regions, we use horizontal flips and RGB
augmentations~\cite{zabari2021inthewild} of $\img$ and obtain a relevancy map
for each augmentation. To reliably detect small objects, we propose a
multi-scale relevancy extractor that \textit{densely} computes relevancies
at different scales and locations in a sliding window fashion (an analogous
relevancy-based approach to \citeauthor{li2022adapting}'s local
attention-pooling). The final relevancy map is averaged across all
augmentations and scales.

Qualitatively, while augmentations help in reducing irrelevant regions (\eg
Fig.~\ref{fig:relevancy_improvements}, from localizing the entire fireplace in
the 1st column to only the vase in the 2nd column), they still often miss small
objects. However, with our multi-scale relevancy extractor, our relevancy maps
can detect small objects such as the tiny ``nintendo switch''
(Fig.~\ref{fig:relevancy_improvements}, bottom right). To support efficient
multi-scale relevancy extraction, we implemented a batch-parallelized ViT
relevancy extractor. At $0.4$ seconds per text label on
average\footnote{measured over 10 repetitions with 100 text labels on a GTX
	1080 Ti machine with a light load. We provide extra details in the appendix.},
we achieve over \textbf{$\times$ 60} speed up compared to the non-parallelized
implementation. With the rise in popularity of ViTs, we believe this efficient
ViT relevancy extractor implementation will be a useful primitive for the
research community and have released it on
\href{https://github.com/columbia-ai-robotics/semantic-abstraction}{Github} and
a \href{https://huggingface.co/spaces/huy-ha/semabs-relevancy}{Hugging Face
	Spaces}.



\subsection{Application to Open Vocabulary Semantic Scene Completion (OVSSC)}
\label{sec:method:ovssc}
\mypara{Task.}  Given as input a single RGB-D image $\img $ and a set of $K$ object class labels, represented as open-vocabulary text $\{\textlabel_k\}^K_{k=1 }$, the task is to complete the 3D geometry of all the partially observed objects referred by $\{\textlabel_k\}^K_{k=1 }$.
In contrast to its closed-world variant~\cite{song2017semantic}, where the object list is fixed for all scenes, OVSSC formulation accounts for context-dependency, which allows the class labels to change from scene to scene and from training to testing.
From this 3D semantic scene completion, robots can use the detailed 3D geometry for low level planning such as grasping and motion planning.

\mypara{Applying SemAbs to OVSSC} is a matter of applying the SemAbs module $K$ times, once per object label.
First, the \firstmodule~abstracts $\img$ and $\{\textlabel_k\}^K_{k=1}$ into $K$ relevancy point clouds $\{\relevancypcd_k\}^K_{k=1}$.
Then, the \secondmodule~extracts $K$ feature volumes $\{\threeDFeatureVolume_k\}^K_{k=1}$, used to decode $K$  3D occupancies $ \{ \mathcal{O}_k\}^K_{k=1}$.
The final OVSSC output is a point-wise argmax of occupancy probability over the $K$ classes.
Query points with the maximum occupancy probability less than a threshold $c$ are labeled as empty.
We provide data generation and training details in the appendix.

\subsection{Application to Visually Obscured Object Localization (VOOL)}
\label{sec:method:vool}

\mypara{Motivation.}
Even after completing partially visible objects in the scene, OVSSC's output may not be complete -- the scene may contain visually obscured or hidden objects (e.g., the ticket on the fireplace in Fig. \ref{fig:method}b).
In such cases, the human user can provide additional hints to the robot using
natural language to help localize the object, which motivates our VOOL task.

\mypara{Task.}
Given as input a single RGB-D image $\img$ of a 3D scene and a description $\mathcal{D}$ of an object's location in the scene, the task is to predict the possible occupancies of the referred object.
Similar to prior works~\cite{liu2021learning}, we assume that the description is in the standard linguistic representation of an object's place~\cite{barbara1993whatwhere}.
Specifically, $\mathcal{D} = \langle \textlabel_\textrm{target},\spatialrelation, \textlabel_\textrm{ref} \rangle$, where the open-vocabulary target object label $\textlabel_\textrm{target}$ (\eg ``rapid test'') is the visually obscured object to be located, the open-vocabulary reference object label $\textlabel_\textrm{ref}$ (\eg ``harry potter book'') is presumably a visible object, and the closed-vocabulary spatial preprosition $\spatialrelation$ describes $\textlabel_\textrm{target}$'s location with respect to $\textlabel_\textrm{ref}$ (\eg ``behind'').
Unlike the referred expression 3D object localization task~\cite{chen2020scanrefer} which assumes full visibility (\ie fused pointcloud), VOOL takes as input only the current RGB-D image, and thus is more suited for a dynamic, partially observable environment.
Further, in contrast to bounding boxes, VOOL's occupancy can be disjoint regions in space, which accounts the multi-modal uncertainty when objects are hidden.
Downstream robotic applications, such as object searching, benefit more from the uncertainty information encoded in a VOOL's output than bounding boxes.

\mypara{Applying SemAbs to VOOL.}
We propose to learn an embedding $\spatialEmbeddingfunction(\spatialrelation)     \in \mathbb{R}^{2D}$ for each spatial relation (Fig.~\ref{fig:method}b, blue box), since the set of spatial prepositions is small and finite~\cite{barbara1993whatwhere}, and use them as follows.
As in OVSSC (\S \ref{sec:method:ovssc}), we can get occupancy feature volumes $\threeDFeatureVolume_\textrm{target}$ and $\threeDFeatureVolume_\textrm{ref}$ for the target and reference objects respectively.
Given a set of fixed query points $Q = \{q_i\}_{i=1}^{N}$, we extract their local feature point clouds $\phi_Q^{\threeDFeatureVolume_\textrm{target}},\phi_Q^{\threeDFeatureVolume_\textrm{ref}} \in {R}^{N\times D}$, then concatenate them to get a new feature pointcloud $\phi_Q^{{\threeDFeatureVolume_\textrm{target}\|\threeDFeatureVolume_\textrm{ref}}} \in {R}^{N\times 2D}$.
Finally, to get the 3D localization occupancy $\mathcal{O}$, we perform a point-wise cosine-similarity between {\small $\phi_Q^{{\threeDFeatureVolume_\textrm{target}\|\threeDFeatureVolume_\textrm{ref}}}$} and $\spatialEmbeddingfunction(\spatialrelation)$.
For this task, we only need to learn the spatial embeddings $\spatialEmbeddingfunction$ conditioned on the occupancy information encoded in $\threeDFeatureVolume_\textrm{target}$ and $\threeDFeatureVolume_\textrm{ref}$, since semantic reasoning of $\textlabel_\textrm{target}$ and $\textlabel_\textrm{ref}$ has been offloaded to the SemAbs~(and, therefore, to CLIP).
We provide data generation and training details in the appendix.
\mysection{Experiments}
\label{sec:eval}

\begin{figure}[t]
	\centering
	\includegraphics[width=\linewidth]{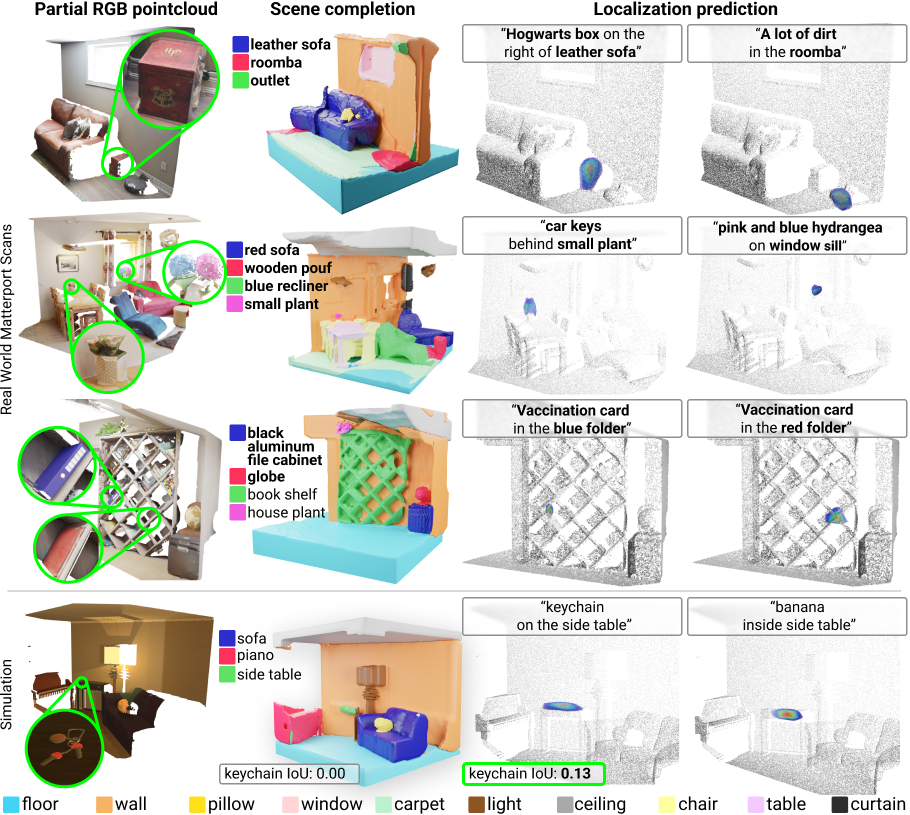}
	\vspace{-3mm}
	\caption{ \footnotesize
		\textbf{Semantic Abstraction inherits CLIP's visual-semantic reasoning skils}
		From distinguishing colors (\eg ``red folder'' v.s. ``blue folder'') to recognizing cultural (\eg ``hogwarts box'') and long-tail  semantic concepts (\eg ``roomba'', ``hydrangea''), our approach offloads such visual-semantic reasoning challenges to CLIP.
		Indoing so, its learned 3D spatial and geometric reasoning skills transfers sim2real in a zero-shot manner.
	} \vspace{-7mm}

	\label{fig:qualitative_results}
\end{figure}

We design experiments to systemically investigate \ours's open-world
generalization abilities to novel\footnote{Here, we define ``novel'' as to
	whether a concept was observed during the 3D training process.}
concepts in 3D scene understanding. Our benchmark include the following
categories: 1) \textbf{Novel Rooms}: we follow AI2-THOR~\cite{ai2thor}'s split,
which holds out 20 rooms for as novel rooms, while training on 100 rooms.
2) \textbf{Novel Visual}: we use AI2-THOR's randomize material and lighting functionality to generate more varied 2D visual conditions of novel rooms.
3) \textbf{Novel Synonyms}: we manually picked out 17 classes which had natural synonyms (\eg ``pillow'' to ``cushion'', complete list included in appendix) and replace their occurrences in text inputs (\ie object label in OVSSC, description in VOOL) to these synonyms.
4) \textbf{Novel Class}: we hold out 6 classes out of the 202 classes in AI2-THOR.
For OVSSC, any view which contains one of these novel classes is held out for testing.
For VOOL, any scene with a description that contains a novel class (as either the target or reference object) is held out.
5) \textbf{Novel Domain}: We also provide qualitative evaluations (Fig.~\ref{fig:qualitative_results}) using real world scans from HM3D~\cite{ramakrishnan2021hm3d}.

\mypara{Metrics \& Baselines.}
For both tasks, we measure the learner's performance with voxel IoU of dimension $32\times32\times32$, where the ground truth of each task is generated as described in \S \ref{sec:method:ovssc} for OVSSC and \S \ref{sec:method:vool} for VOOL.
We compare with following categories of baselines:
\begin{itemize}[leftmargin=6mm]\setlength\itemsep{-0.0em}
	\vspace{-2mm}\item \textbf{Semantic-Aware (SemAware).}
	      To evaluate the effectiveness of semantic abstraction, we design baselines that perform visual-semantic reasoning themselves by taking RGB point clouds as input (instead of relevancy point clouds).
	      For OVSSC, the baseline's feature point cloud is supervised with BCE to be aligned (\ie using cosine similarity) with the correct object class text embedding.
	      Similarly, for VOOL, the feature point cloud's cosine similarity with an embedding of the entire localization description (\eg ``N95s inside the trash'') is supervised using BCE to the target object occupancy.

	      \vspace{-1mm}\item \textbf{Semantic \& Spatial Abstraction (ClipSpatial).}
	      What if we off-load both 3D and visual-semantic reasoning to 2D VLMs (instead
	      of just the latter, as in our approach)? To answer this, we designed
	      CLIPSpatial, a VOOL baseline that uses relevancy maps for the entire
	      description (\eg ``N95s inside the trash'') as input. We expect this approach
	      to perform poorly since it has been demonstrated that current 2D VLMs
	      \cite{radford2021learning,li2021align} struggles with spatial
	      relations~\cite{subramanian2022reclip,gadre2022clip,ramesh2022hierarchical,yuksekgonul2022visionlanguage}.
	      \vspace{-1mm}\item \textbf{Naive Relevancy Extraction. (SemAbs + \cite{chefer2021generic})}
	      To investigate the effects of relevancy quality, we replace our multi-scale relevancy extractor with \citeauthor{chefer2021generic}~\cite{chefer2021generic}'s approach.

\end{itemize}
\vspace{-3mm}
The first two baselines can be seen as two extremes on a spectrum of how much
we rely on pretrained 2D VLMs. While SemAware learns both semantics and 3D
spatial reasoning, CLIPSpatial delegates both to pretrained 2D VLMs. Our
approach is in between these two extremes, designed such that it builds on 2D
VLMs' visual robustness and open-classification strengths while addressing its
spatial reasoning weaknesses.

\mysubsection{Open-world Evaluation Results}
Tab. \ref{tab:ovssc} and \ref{tab:vool} summarize the quantitative results, and Fig. \ref{fig:qualitative_results} shows real world qualitative results tested on Matterport scenes~\cite{ramakrishnan2021hm3d}.
More results and comparisons can be found on the \href{https://semantic-abstraction.cs.columbia.edu/}{project website}.

\mypara{Semantic Abstraction simplifies open-world generalization.}
SemAware baselines especially struggle to generalize to novel semantic classes (Tab.~\ref{tab:ovssc}, Tab.~\ref{tab:vool}).
\begin{wraptable}{r}{0.46\textwidth}
	\addtolength{\tabcolsep}{-4pt}
	\setlength{\tabcolsep}{0.07cm}
	\footnotesize
	\centering
	\begin{tabular}{lcccc}
		\toprule
		\multirow{2}{*}{Approach}       & \multicolumn{4}{c}{\textbf{Novel}}                                                        \\\cmidrule(lr){2-5}
		                                & \textbf{Room}                      & \textbf{Visual} & \textbf{Synonyms} & \textbf{Class} \\\midrule
		SemAware                        & 32.2                               & 31.9            & 20.2              & 0.0            \\
		SemAbs+\cite{chefer2021generic} & 26.6                               & 24.3            & 17.8              & 12.2           \\
		\midrule
		Ours                            & \textbf{40.1}                      & \textbf{36.4}   & \textbf{33.4}     & \textbf{37.9}  \\
		\bottomrule
	\end{tabular}
	\vspace{-1mm}
	\caption{ \footnotesize
		\centering
		\textbf{Open-Vocab Semantic Scene Completion}
	}
	\vspace{-5mm}
	\label{tab:ovssc}
\end{wraptable}
We hypothesize that observing text embeddings during training causes these baselines to specialize to this distribution of embeddings, such that some synonyms and most novel class labels are out-of-distribution inputs.
Our approach significantly outperforms the SemAware baseline across the board, suggesting that \ours~not only simplifies open-world generalization but is also a strong inductive bias for learning the task.

\mypara{Semantic \& Spatial Abstraction generalizes poorly.}
The CLIPSpatial baseline achieves 8.0-9.0 IoU worse than our approach in all generalization scenarios (Tab.~\ref{tab:vool}).
This indicates that our design of learning spatial reasoning instead of relying on CLIP was crucial to semantic abstraction's performance.

\begin{wraptable}{r}{0.44\textwidth}
	\vspace{-2mm}
	\addtolength{\tabcolsep}{-4pt}
	\setlength{\tabcolsep}{0.07cm}
	\footnotesize
	\centering
	\begin{tabular}{lcccc}
		\toprule
		\multirow{2}{*}{Approach}       & \multicolumn{4}{c}{\textbf{Novel}}                                                        \\\cmidrule(lr){2-5}
		                                & \textbf{Room}                      & \textbf{Visual} & \textbf{Synonyms} & \textbf{Class} \\\midrule
		SemAware                        & 12.1                               & 11.9            & 11.4              & 3.0            \\
		SemAbs+\cite{chefer2021generic} & 9.1                                & 8.8             & 10.7              & 4.0            \\
		CLIPSpatial                     & 11.7                               & 10.1            & 14.3              & 11.2           \\
		\midrule
		Ours                            & \textbf{20.9}                      & \textbf{19.2}   & \textbf{23.4}     & \textbf{19.7}  \\
		\bottomrule
	\end{tabular}
	\caption{\footnotesize \centering{\textbf{Visually Obscured Object Localization}}}
	\label{tab:vool}
	\vspace{-6mm}
\end{wraptable}
\mypara{Semantic Abstraction requires high quality relevancy maps.}
Intuitively, the model can only output high quality completions if its relevancy maps are also high quality (\ie highlights the target object, especially when the object is small).
Our average IoU is $\mathbf{\times 1.5}$ (novel room, visual, synonyms, Table~\ref{tab:ovssc}),
over $\mathbf{\times 2.0}$ (novel classes, Table~\ref{tab:ovssc}, novel room, visual, synonyms, Table~\ref{tab:vool}) or $\mathbf{\times 5.0}$ (novel class, Table~\ref{tab:vool}) compared to SemAbs + \cite{chefer2021generic}, which demonstrates the importance of our relevancy extractor in a successful application of \ours.

\mypara{Inheriting VLM zero-shot robustness enables Sim2Real transfer.}
Despite training in simulation, our approach can perform completion and localization from real-world matterport scans~\cite{ramakrishnan2021hm3d} (Fig.~\ref{fig:teaser}  and \ref{fig:qualitative_results}).
From novel vocabulary for describing materials (\eg ``leather sofa'', ``wooden pouf''), colors (\eg ``red folder'', ``blue folder''), conjunctions of them (\eg ``black aluminum file cabinet'') to novel semantic classes (\eg ``roomba'', ``Hogwarts box''), our approach unlocks CLIP's visual robustness and large repertoire of visual-semantic concepts to 3D scene understanding.
When OVSSC's completion misses objects (\eg ``hydrangea'', ``keychain''), VOOL can still localize them (\eg +0.13 IoU for keychain).

\begin{figure}
	\centering
	\includegraphics[width=\linewidth]{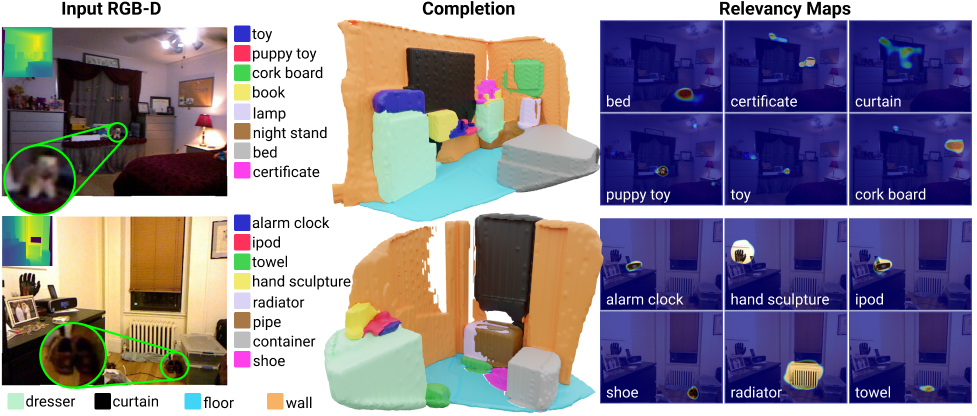}
	\caption{
		\textbf{Qualitative result on NYUCAD~\cite{silberman2012indoor,guo2015predicting} dataset.}
		Our model can distinguish between ``puppy toy'', ``ipod'', ``certificate'', and ``hand sculpture'',
		while all prior works would've output ``obj''.
	}
	\vspace{-2mm}
	\label{fig:nyucad}
\end{figure}

\mysubsection{Zero-shot Evaluation Results}

The NYUv2 CAD dataset~\cite{silberman2012indoor, guo2015predicting} is a popular closed-vocabulary SSC benchmark.
Although it contains 894 object classes, prior works~\cite{song2017semantic,cai2021semantic,zhang2019cascaded} collapse it down to 11 classes to avoid the challenge in learning the ``long-tail'' categories (\eg,``puppy toy'',``ipod'', ``hand sculture'').
We evaluate our model zero-shot on NYUv2 directly on \textbf{all 894 classes} as well as the 11 classes for comparison (Fig.~\ref{fig:nyucad}, Tab.~\ref{tab:nyucad}, Ours).
We also train a SemAbs model from scratch in NYU to investigate the generalization gap between our dataset and NYUv2 (Tab.~\ref{tab:nyucad}, Ours NYU).
We use the same training/testing split~\cite{song2017semantic}.
Similar to prior works~\cite{song2017semantic}, we report volumetric IoU at a voxel dimension of 60.
As in our THOR dataset, we only train $f_{encode}$ and $f_{decode}$ in the semantic abstracted 3D module, not the 2D VLM.
To our knowledge, our approach is the first to output the full list of 894 object categories in this benchmark.

\mypara{Open-world versus Closed-world formulation.}
Our work investigates open-world 3D scene understanding, which means specializing to one data distribution is not our focus. However, we hope these results on a standard closed-world benchmark along with our discussion below are helpful to the research community in understanding practical trade-offs in algorithm and evaluation design between the open-world versus closed-world formulation. Below summarizes our key findings:
\begin{itemize}[leftmargin=4mm]\setlength\itemsep{-0.1em}
	\vspace{-2mm}
	\item[+] \textbf{Flexibility to object labels.}
		Our framework directly outputs the full 894 object categories, of which 108 do not appear in the training split, and its predictions can map back to the collapsed 11 classes.
		This has never been attempted by prior works.
		This flexibility is achieved by building on top of 2D VLMs.

	\item[+] \textbf{Robust detection of small, long-tail object categories.}
		``Due to the heavy-tailed distribution of object instances in natural
		images, a large number of objects will occur too infrequently to build usable
		recognition models''~\cite{gupta2010blocks}.
		However, our approach can robustly detect not just long-tail object classes, but also small ones -- challenging scenarios for perception systems.
		For instance, from Fig.~\ref{fig:nyucad}, our multi-scale relevancy extractor can output highly localized relevancy maps around ``puppy toy'',``ipod'', and ``hand sculpture''.

	\item[-] \textbf{Inability to finetune large pretrained models.}
		As shown in prior works~\cite{radford2021learning,wortsman2021robust,pham2021combined,andreassen2021evolution,zhou2021denseclip}, directly finetuning 2D VLMs on small datasets hurts their robustness. We cannot finetune the visual encoders (and thus, its relevancy activations) to specialize on the visual and class distribution on a small dataset (e.g., NYUv2) without sacrificing it robustness and generality.
		This is an open and important research direction and recent work~\cite{soup2022wortsman} has demonstrated promising results in image classification.

	\item[-] \textbf{Bias from image caption data.}
		The typical training paradigm of current 2D VLMs~\cite{radford2021learning,pham2021combined,jia2021scaling} is contrastive learning over image-caption pairs on the internet.
		While their dataset covers a large number of semantic concepts, they still have their limits. 
		For instance, we observed that for ``wall'' and ``ceiling'' -- two classes which closed-world approaches typically achieve high performance for -- our approach achieved an much lower IoU of only $33.9$ and $22.6$.
		Since these classes are often not the subject of internet captions, our approach inherit this bias from 2D VLMs.
		However, our framework is \textbf{VLM-agnostic}, which means it can continue to benefit from future and more powerful versions of 2D VLMs.
		\vspace{-2mm}
\end{itemize}

\begin{table}
	\footnotesize
	\begin{tabular}{l|cccccccccccc}
		\toprule
		Approach                       & ceil. & floor & wall & win. & chair & bed  & sofa & table & tvs  & furn. & objs. & avg. \\
		\midrule
		SSCNet~\cite{song2017semantic} & 32.5  & 92.6  & 40.2 & 8.9  & 33.9  & 57   & 59.5 & 28.3  & 8.1  & 44.8  & 25.1  & 36.6 \\
		SISNet~\cite{cai2021semantic}  & 63.4  & 94.4  & 67.2 & 52.4 & 59.2  & 77.9 & 71.1 & 51.8  & 46.2 & 65.8  & 48.8  & 63.5 \\
		Ours \textbf{Supervised}       & 22.6  & 46.1  & 33.9 & 35.9 & 23.9  & 55.9 & 37.9 & 19.7  & 30.8 & 39.8  & 27.7  & 34.0 \\
		\midrule
		Ours \textbf{Zeroshot}         & 13.7  & 17.3  & 13.5 & 25.2 & 15.2  & 33.3 & 31.5 & 12.0  & 23.7 & 25.6  & 19.9  & 21.0 \\
		\bottomrule
	\end{tabular}
	\vspace{1mm}
	\caption{\textbf{NYUv2 CAD Semantic Scene Completion.}}
	\vspace{-9mm}
	\label{tab:nyucad}
\end{table}

\mysubsection{Limitations and assumptions}

Our key assumption is that geometry and spatial reasoning can be done in a semantic agnostic manner. This assumption does not hold when we care about completing detailed object geometry, which is often semantic-dependent. However, we believe that coarse 3D reasoning with paired semantic labels may already useful for open-world robotic applications such as object retrieval. Second, while we have assumed a simple description syntax and a small set of spatial prepositions for object localization, incorporating natural language understanding (\ie to support open-set phrases for richer spatial descriptions) into open-world 3D scene understanding is an important extension of our work. Lastly, we assumed that relevancy maps highlight only relevant regions. Limited by current VLMs, our relevancy extractor can sometimes fail to distinguish highly related concepts (\eg book v.s. bookshelf). However, since \ours~is VLM-agnostic, this limitation could be addressed by using other VLMs.
\mysection{Conclusion}
We proposed Semantic Abstraction, a framework that equips 2D VLMs with 3D spatial capabilities for open-world 3D scene understanding tasks.
By abstracting RGB and semantic label inputs as relevancy maps, our learned 3D spatial and geometric reasoning skills generalizes to novel vocabulary for object attributes and nouns, visual properties, semantic classes, and domains.
Since scene completion and 3D object localization are fundamental primitives for many robotic tasks, we believe our approach can assist in bridging currently close-world robotics tasks, such as object retrieval, into the open-world formulation. More importantly, we hope our framework encourages roboticists to embrace building systems around large pretrained models, as a path towards open-world robotics.

\mypara{Acknowledgments:} We would like to thank  Samir Yitzhak Gadre, Cheng Chi and Zhenjia Xu for their helpful feedback and fruitful discussions. This work was supported in part by NSF Award \#2143601, \#2132519 JP Morgan Faculty Research Award, and Google Research Award. The views and conclusions contained herein are those of the authors and should not be interpreted as necessarily representing the official policies, either expressed or implied, of the sponsors.

\bibliography{references}

\begin{thebibliography}{53}
\providecommand{\natexlab}[1]{#1}
\providecommand{\url}[1]{\texttt{#1}}
\expandafter\ifx\csname urlstyle\endcsname\relax
  \providecommand{\doi}[1]{doi: #1}\else
  \providecommand{\doi}{doi: \begingroup \urlstyle{rm}\Url}\fi

\bibitem[Kolve et~al.(2017)Kolve, Mottaghi, Han, VanderBilt, Weihs, Herrasti,
  Gordon, Zhu, Gupta, and Farhadi]{ai2thor}
E.~Kolve, R.~Mottaghi, W.~Han, E.~VanderBilt, L.~Weihs, A.~Herrasti, D.~Gordon,
  Y.~Zhu, A.~Gupta, and A.~Farhadi.
\newblock {AI2-THOR: An Interactive 3D Environment for Visual AI}.
\newblock \emph{arXiv}, 2017.

\bibitem[Song et~al.(2015)Song, Lichtenberg, and Xiao]{song2015sun}
S.~Song, S.~P. Lichtenberg, and J.~Xiao.
\newblock Sun rgb-d: A rgb-d scene understanding benchmark suite.
\newblock In \emph{Proceedings of the IEEE conference on computer vision and
  pattern recognition}, pages 567--576, 2015.

\bibitem[Chang et~al.(2015)Chang, Funkhouser, Guibas, Hanrahan, Huang, Li,
  Savarese, Savva, Song, Su, et~al.]{chang2015shapenet}
A.~X. Chang, T.~Funkhouser, L.~Guibas, P.~Hanrahan, Q.~Huang, Z.~Li,
  S.~Savarese, M.~Savva, S.~Song, H.~Su, et~al.
\newblock Shapenet: An information-rich 3d model repository.
\newblock \emph{arXiv preprint arXiv:1512.03012}, 2015.

\bibitem[Ramakrishnan et~al.(2021)Ramakrishnan, Gokaslan, Wijmans, Maksymets,
  Clegg, Turner, Undersander, Galuba, Westbury, Chang, Savva, Zhao, and
  Batra]{ramakrishnan2021hm3d}
S.~K. Ramakrishnan, A.~Gokaslan, E.~Wijmans, O.~Maksymets, A.~Clegg, J.~M.
  Turner, E.~Undersander, W.~Galuba, A.~Westbury, A.~X. Chang, M.~Savva,
  Y.~Zhao, and D.~Batra.
\newblock Habitat-matterport 3d dataset ({HM}3d): 1000 large-scale 3d
  environments for embodied {AI}.
\newblock In \emph{Thirty-fifth Conference on Neural Information Processing
  Systems Datasets and Benchmarks Track (Round 2)}, 2021.
\newblock URL \url{https://openreview.net/forum?id=-v4OuqNs5P}.

\bibitem[Szot et~al.(2021)Szot, Clegg, Undersander, Wijmans, Zhao, Turner,
  Maestre, Mukadam, Chaplot, Maksymets, Gokaslan, Vondrus, Dharur, Meier,
  Galuba, Chang, Kira, Koltun, Malik, Savva, and Batra]{szot2021habitat}
A.~Szot, A.~Clegg, E.~Undersander, E.~Wijmans, Y.~Zhao, J.~Turner, N.~Maestre,
  M.~Mukadam, D.~Chaplot, O.~Maksymets, A.~Gokaslan, V.~Vondrus, S.~Dharur,
  F.~Meier, W.~Galuba, A.~Chang, Z.~Kira, V.~Koltun, J.~Malik, M.~Savva, and
  D.~Batra.
\newblock Habitat 2.0: Training home assistants to rearrange their habitat.
\newblock In \emph{Advances in Neural Information Processing Systems
  (NeurIPS)}, 2021.

\bibitem[Chen et~al.(2020)Chen, Chang, and Nie{\ss}ner]{chen2020scanrefer}
D.~Z. Chen, A.~X. Chang, and M.~Nie{\ss}ner.
\newblock Scanrefer: 3d object localization in rgb-d scans using natural
  language.
\newblock In \emph{European Conference on Computer Vision}, pages 202--221.
  Springer, 2020.

\bibitem[Radford et~al.(2021)Radford, Kim, Hallacy, Ramesh, Goh, Agarwal,
  Sastry, Askell, Mishkin, Clark, et~al.]{radford2021learning}
A.~Radford, J.~W. Kim, C.~Hallacy, A.~Ramesh, G.~Goh, S.~Agarwal, G.~Sastry,
  A.~Askell, P.~Mishkin, J.~Clark, et~al.
\newblock Learning transferable visual models from natural language
  supervision.
\newblock In \emph{International Conference on Machine Learning}, pages
  8748--8763. PMLR, 2021.

\bibitem[Li et~al.(2021)Li, Selvaraju, Gotmare, Joty, Xiong, and
  Hoi]{li2021align}
J.~Li, R.~Selvaraju, A.~Gotmare, S.~Joty, C.~Xiong, and S.~C.~H. Hoi.
\newblock Align before fuse: Vision and language representation learning with
  momentum distillation.
\newblock \emph{Advances in Neural Information Processing Systems}, 34, 2021.

\bibitem[Alayrac et~al.(2022)Alayrac, Donahue, Luc, Miech, Barr, Hasson, Lenc,
  Mensch, Millican, Reynolds, et~al.]{alayrac2022flamingo}
J.-B. Alayrac, J.~Donahue, P.~Luc, A.~Miech, I.~Barr, Y.~Hasson, K.~Lenc,
  A.~Mensch, K.~Millican, M.~Reynolds, et~al.
\newblock Flamingo: a visual language model for few-shot learning.
\newblock \emph{arXiv preprint arXiv:2204.14198}, 2022.

\bibitem[Pham et~al.(2021)Pham, Dai, Ghiasi, Liu, Yu, Luong, Tan, and
  Le]{pham2021combined}
H.~Pham, Z.~Dai, G.~Ghiasi, H.~Liu, A.~W. Yu, M.-T. Luong, M.~Tan, and Q.~V.
  Le.
\newblock Combined scaling for zero-shot transfer learning.
\newblock \emph{arXiv preprint arXiv:2111.10050}, 2021.

\bibitem[Jia et~al.(2021)Jia, Yang, Xia, Chen, Parekh, Pham, Le, Sung, Li, and
  Duerig]{jia2021scaling}
C.~Jia, Y.~Yang, Y.~Xia, Y.-T. Chen, Z.~Parekh, H.~Pham, Q.~Le, Y.-H. Sung,
  Z.~Li, and T.~Duerig.
\newblock Scaling up visual and vision-language representation learning with
  noisy text supervision.
\newblock In \emph{International Conference on Machine Learning}, pages
  4904--4916. PMLR, 2021.

\bibitem[Wortsman et~al.(2021)Wortsman, Ilharco, Li, Kim, Hajishirzi, Farhadi,
  Namkoong, and Schmidt]{wortsman2021robust}
M.~Wortsman, G.~Ilharco, M.~Li, J.~W. Kim, H.~Hajishirzi, A.~Farhadi,
  H.~Namkoong, and L.~Schmidt.
\newblock Robust fine-tuning of zero-shot models.
\newblock \emph{arXiv preprint arXiv:2109.01903}, 2021.

\bibitem[Andreassen et~al.(2021)Andreassen, Bahri, Neyshabur, and
  Roelofs]{andreassen2021evolution}
A.~Andreassen, Y.~Bahri, B.~Neyshabur, and R.~Roelofs.
\newblock The evolution of out-of-distribution robustness throughout
  fine-tuning.
\newblock \emph{arXiv preprint arXiv:2106.15831}, 2021.

\bibitem[Subramanian et~al.(2022)Subramanian, Merrill, Darrell, Gardner, Singh,
  and Rohrbach]{subramanian2022reclip}
S.~Subramanian, W.~Merrill, T.~Darrell, M.~Gardner, S.~Singh, and A.~Rohrbach.
\newblock Reclip: A strong zero-shot baseline for referring expression
  comprehension.
\newblock \emph{arXiv preprint arXiv:2204.05991}, 2022.

\bibitem[Gadre et~al.(2022)Gadre, Wortsman, Ilharco, Schmidt, and
  Song]{gadre2022clip}
S.~Y. Gadre, M.~Wortsman, G.~Ilharco, L.~Schmidt, and S.~Song.
\newblock Clip on wheels: Zero-shot object navigation as object localization
  and exploration.
\newblock \emph{arXiv preprint arXiv:2203.10421}, 2022.

\bibitem[Ramesh et~al.(2022)Ramesh, Dhariwal, Nichol, Chu, and
  Chen]{ramesh2022hierarchical}
A.~Ramesh, P.~Dhariwal, A.~Nichol, C.~Chu, and M.~Chen.
\newblock Hierarchical text-conditional image generation with clip latents.
\newblock \emph{arXiv preprint arXiv:2204.06125}, 2022.

\bibitem[Dosovitskiy et~al.(2020)Dosovitskiy, Beyer, Kolesnikov, Weissenborn,
  Zhai, Unterthiner, Dehghani, Minderer, Heigold, Gelly,
  et~al.]{dosovitskiy2020image}
A.~Dosovitskiy, L.~Beyer, A.~Kolesnikov, D.~Weissenborn, X.~Zhai,
  T.~Unterthiner, M.~Dehghani, M.~Minderer, G.~Heigold, S.~Gelly, et~al.
\newblock An image is worth 16x16 words: Transformers for image recognition at
  scale.
\newblock \emph{arXiv preprint arXiv:2010.11929}, 2020.

\bibitem[Chefer et~al.(2021)Chefer, Gur, and Wolf]{chefer2021generic}
H.~Chefer, S.~Gur, and L.~Wolf.
\newblock Generic attention-model explainability for interpreting bi-modal and
  encoder-decoder transformers.
\newblock In \emph{Proceedings of the IEEE/CVF International Conference on
  Computer Vision}, pages 397--406, 2021.

\bibitem[Zhou et~al.(2021)Zhou, Loy, and Dai]{zhou2021denseclip}
C.~Zhou, C.~C. Loy, and B.~Dai.
\newblock Denseclip: Extract free dense labels from clip.
\newblock \emph{arXiv preprint arXiv:2112.01071}, 2021.

\bibitem[Rao et~al.(2022)Rao, Zhao, Chen, Tang, Zhu, Huang, Zhou, and
  Lu]{rao2022denseclip}
Y.~Rao, W.~Zhao, G.~Chen, Y.~Tang, Z.~Zhu, G.~Huang, J.~Zhou, and J.~Lu.
\newblock Denseclip: Language-guided dense prediction with context-aware
  prompting.
\newblock In \emph{Proceedings of the IEEE/CVF Conference on Computer Vision
  and Pattern Recognition}, pages 18082--18091, 2022.

\bibitem[Li et~al.(2022)Li, Shakhnarovich, and Yeh]{li2022adapting}
J.~Li, G.~Shakhnarovich, and R.~A. Yeh.
\newblock Adapting clip for phrase localization without further training.
\newblock \emph{arXiv preprint arXiv:2204.03647}, 2022.

\bibitem[Selvaraju et~al.(2017)Selvaraju, Cogswell, Das, Vedantam, Parikh, and
  Batra]{selvaraju2017grad}
R.~R. Selvaraju, M.~Cogswell, A.~Das, R.~Vedantam, D.~Parikh, and D.~Batra.
\newblock Grad-cam: Visual explanations from deep networks via gradient-based
  localization.
\newblock In \emph{Proceedings of the IEEE international conference on computer
  vision}, pages 618--626, 2017.

\bibitem[Mokady et~al.(2021)Mokady, Hertz, and Bermano]{mokady2021clipcap}
R.~Mokady, A.~Hertz, and A.~H. Bermano.
\newblock Clipcap: Clip prefix for image captioning.
\newblock \emph{arXiv preprint arXiv:2111.09734}, 2021.

\bibitem[Song et~al.(2022)Song, Dong, Zhang, Liu, and Wei]{song2022clip}
H.~Song, L.~Dong, W.-N. Zhang, T.~Liu, and F.~Wei.
\newblock Clip models are few-shot learners: Empirical studies on vqa and
  visual entailment.
\newblock \emph{arXiv preprint arXiv:2203.07190}, 2022.

\bibitem[Gao et~al.(2021)Gao, Liu, Chen, Chang, Zhang, and
  Yuan]{gao2021clip2tv}
Z.~Gao, J.~Liu, S.~Chen, D.~Chang, H.~Zhang, and J.~Yuan.
\newblock Clip2tv: An empirical study on transformer-based methods for
  video-text retrieval.
\newblock \emph{arXiv preprint arXiv:2111.05610}, 2021.

\bibitem[Wang et~al.(2021)Wang, Xing, and Liu]{wang2021actionclip}
M.~Wang, J.~Xing, and Y.~Liu.
\newblock Actionclip: A new paradigm for video action recognition.
\newblock \emph{arXiv preprint arXiv:2109.08472}, 2021.

\bibitem[Zabari and Hoshen(2021)]{zabari2021inthewild}
N.~Zabari and Y.~Hoshen.
\newblock Semantic segmentation in-the-wild without seeing any segmentation
  examples.
\newblock \emph{CoRR}, abs/2112.03185, 2021.
\newblock URL \url{https://arxiv.org/abs/2112.03185}.

\bibitem[Shridhar et~al.(2021)Shridhar, Manuelli, and Fox]{shridhar2021cliport}
M.~Shridhar, L.~Manuelli, and D.~Fox.
\newblock Cliport: What and where pathways for robotic manipulation.
\newblock In \emph{Proceedings of the 5th Conference on Robot Learning (CoRL)},
  2021.

\bibitem[Zeng et~al.(2022)Zeng, Wong, Welker, Choromanski, Tombari, Purohit,
  Ryoo, Sindhwani, Lee, Vanhoucke, et~al.]{zeng2022socratic}
A.~Zeng, A.~Wong, S.~Welker, K.~Choromanski, F.~Tombari, A.~Purohit, M.~Ryoo,
  V.~Sindhwani, J.~Lee, V.~Vanhoucke, et~al.
\newblock Socratic models: Composing zero-shot multimodal reasoning with
  language.
\newblock \emph{arXiv preprint arXiv:2204.00598}, 2022.

\bibitem[{Shuran Song} and Xiao(2014)]{SlidingShapes}
{Shuran Song} and J.~Xiao.
\newblock {Sliding Shapes} for {3D} object detection in depth images.
\newblock 2014.

\bibitem[{Shuran Song} and Xiao(2016)]{DSS}
{Shuran Song} and J.~Xiao.
\newblock Deep sliding shapes for amodal {3D} object detection in rgb-d images.
\newblock 2016.

\bibitem[Hou et~al.(2019)Hou, Dai, and Nie{\ss}ner]{hou20193dsis}
J.~Hou, A.~Dai, and M.~Nie{\ss}ner.
\newblock 3d-sis: 3d semantic instance segmentation of rgb-d scans.
\newblock In \emph{Proc. Computer Vision and Pattern Recognition (CVPR), IEEE},
  2019.

\bibitem[Song et~al.(2017)Song, Yu, Zeng, Chang, Savva, and
  Funkhouser]{song2017semantic}
S.~Song, F.~Yu, A.~Zeng, A.~X. Chang, M.~Savva, and T.~Funkhouser.
\newblock Semantic scene completion from a single depth image.
\newblock In \emph{Proceedings of the IEEE conference on computer vision and
  pattern recognition}, pages 1746--1754, 2017.

\bibitem[Dai et~al.(2018)Dai, Ritchie, Bokeloh, Reed, Sturm, and
  Nie{\ss}ner]{dai2018scancomplete}
A.~Dai, D.~Ritchie, M.~Bokeloh, S.~Reed, J.~Sturm, and M.~Nie{\ss}ner.
\newblock Scancomplete: Large-scale scene completion and semantic segmentation
  for 3d scans.
\newblock In \emph{Proceedings of the IEEE Conference on Computer Vision and
  Pattern Recognition}, pages 4578--4587, 2018.

\bibitem[Avetisyan et~al.(2019)Avetisyan, Dahnert, Dai, Savva, Chang, and
  Nie{\ss}ner]{avetisyan2019scan2cad}
A.~Avetisyan, M.~Dahnert, A.~Dai, M.~Savva, A.~X. Chang, and M.~Nie{\ss}ner.
\newblock Scan2cad: Learning cad model alignment in rgb-d scans.
\newblock In \emph{Proceedings of the IEEE/CVF Conference on computer vision
  and pattern recognition}, pages 2614--2623, 2019.

\bibitem[Roh et~al.(2022)Roh, Desingh, Farhadi, and Fox]{roh2022languagerefer}
J.~Roh, K.~Desingh, A.~Farhadi, and D.~Fox.
\newblock Languagerefer: Spatial-language model for 3d visual grounding.
\newblock In \emph{Conference on Robot Learning}, pages 1046--1056. PMLR, 2022.

\bibitem[Barbara and Jackendoff(1993)]{barbara1993whatwhere}
L.~Barbara and R.~Jackendoff.
\newblock “what” and “where” in spatial language and spatial cognition.
\newblock \emph{Behavioral and brain sciences}, 16:\penalty0 217--238, 1993.

\bibitem[Hayward and Tarr(1995)]{hayward1995spatial}
W.~G. Hayward and M.~J. Tarr.
\newblock Spatial language and spatial representation.
\newblock \emph{Cognition}, 55\penalty0 (1):\penalty0 39--84, 1995.

\bibitem[Tulsiani et~al.(2017)Tulsiani, Su, Guibas, Efros, and
  Malik]{tulsiani2017learning}
S.~Tulsiani, H.~Su, L.~J. Guibas, A.~A. Efros, and J.~Malik.
\newblock Learning shape abstractions by assembling volumetric primitives.
\newblock In \emph{Proceedings of the IEEE Conference on Computer Vision and
  Pattern Recognition}, pages 2635--2643, 2017.

\bibitem[Binford(1971)]{binford1971visual}
I.~Binford.
\newblock Visual perception by computer.
\newblock In \emph{IEEE Conference of Systems and Control}, 1971.

\bibitem[Gupta et~al.(2010)Gupta, Efros, and Hebert]{gupta2010blocks}
A.~Gupta, A.~A. Efros, and M.~Hebert.
\newblock Blocks world revisited: Image understanding using qualitative
  geometry and mechanics.
\newblock In \emph{European Conference on Computer Vision}, pages 482--496.
  Springer, 2010.

\bibitem[{\c{C}}i{\c{c}}ek et~al.(2016){\c{C}}i{\c{c}}ek, Abdulkadir, Lienkamp,
  Brox, and Ronneberger]{abdulkadir2016unet3d}
{\"{O}}.~{\c{C}}i{\c{c}}ek, A.~Abdulkadir, S.~S. Lienkamp, T.~Brox, and
  O.~Ronneberger.
\newblock 3d u-net: Learning dense volumetric segmentation from sparse
  annotation.
\newblock \emph{CoRR}, abs/1606.06650, 2016.
\newblock URL \url{http://arxiv.org/abs/1606.06650}.

\bibitem[Chefer et~al.(2022)Chefer, Schwartz, and Wolf]{chefer2022optimizing}
H.~Chefer, I.~Schwartz, and L.~Wolf.
\newblock Optimizing relevance maps of vision transformers improves robustness.
\newblock In \emph{Thirty-Sixth Conference on Neural Information Processing
  Systems}, 2022.
\newblock URL \url{https://openreview.net/forum?id=upuYKQiyxa_}.

\bibitem[Chefer et~al.(2021)Chefer, Gur, and Wolf]{chefer2021transformer}
H.~Chefer, S.~Gur, and L.~Wolf.
\newblock Transformer interpretability beyond attention visualization.
\newblock In \emph{Proceedings of the IEEE/CVF Conference on Computer Vision
  and Pattern Recognition}, pages 782--791, 2021.

\bibitem[Liu et~al.(2022)Liu, Li, Guo, Kong, Li, and Wang]{liu2022rethinking}
Y.~Liu, H.~Li, Y.~Guo, C.~Kong, J.~Li, and S.~Wang.
\newblock Rethinking attention-model explainability through faithfulness
  violation test.
\newblock \emph{arXiv preprint arXiv:2201.12114}, 2022.

\bibitem[Shen et~al.(2021)Shen, Li, Tan, Bansal, Rohrbach, Chang, Yao, and
  Keutzer]{shen2021much}
S.~Shen, L.~H. Li, H.~Tan, M.~Bansal, A.~Rohrbach, K.-W. Chang, Z.~Yao, and
  K.~Keutzer.
\newblock How much can clip benefit vision-and-language tasks?
\newblock \emph{arXiv preprint arXiv:2107.06383}, 2021.

\bibitem[Liu et~al.(2021)Liu, Li, Du, Tenenbaum, and Torralba]{liu2021learning}
N.~Liu, S.~Li, Y.~Du, J.~Tenenbaum, and A.~Torralba.
\newblock Learning to compose visual relations.
\newblock \emph{Advances in Neural Information Processing Systems}, 34, 2021.

\bibitem[Yuksekgonul et~al.(2022)Yuksekgonul, Bianchi, Kalluri, Jurafsky, and
  Zou]{yuksekgonul2022visionlanguage}
M.~Yuksekgonul, F.~Bianchi, P.~Kalluri, D.~Jurafsky, and J.~Zou.
\newblock When and why vision-language models behave like bags-of-words, and
  what to do about it?, 2022.

\bibitem[Silberman et~al.(2012)Silberman, Hoiem, Kohli, and
  Fergus]{silberman2012indoor}
N.~Silberman, D.~Hoiem, P.~Kohli, and R.~Fergus.
\newblock Indoor segmentation and support inference from rgbd images.
\newblock In \emph{European conference on computer vision}, pages 746--760.
  Springer, 2012.

\bibitem[Guo et~al.(2015)Guo, Zou, and Hoiem]{guo2015predicting}
R.~Guo, C.~Zou, and D.~Hoiem.
\newblock Predicting complete 3d models of indoor scenes.
\newblock \emph{arXiv preprint arXiv:1504.02437}, 2015.

\bibitem[Cai et~al.(2021)Cai, Chen, Zhang, Lin, Wang, and Li]{cai2021semantic}
Y.~Cai, X.~Chen, C.~Zhang, K.-Y. Lin, X.~Wang, and H.~Li.
\newblock Semantic scene completion via integrating instances and scene
  in-the-loop.
\newblock In \emph{Proceedings of the IEEE/CVF Conference on Computer Vision
  and Pattern Recognition}, pages 324--333, 2021.

\bibitem[Zhang et~al.(2019)Zhang, Liu, Lei, Lu, and Yang]{zhang2019cascaded}
P.~Zhang, W.~Liu, Y.~Lei, H.~Lu, and X.~Yang.
\newblock Cascaded context pyramid for full-resolution 3d semantic scene
  completion.
\newblock In \emph{Proceedings of the IEEE/CVF International Conference on
  Computer Vision}, pages 7801--7810, 2019.

\bibitem[Wortsman et~al.(2022)Wortsman, Ilharco, Gadre, Roelofs, Gontijo-Lopes,
  Morcos, Namkoong, Farhadi, Carmon, Kornblith, and Schmidt]{soup2022wortsman}
M.~Wortsman, G.~Ilharco, S.~Y. Gadre, R.~Roelofs, R.~Gontijo-Lopes, A.~S.
  Morcos, H.~Namkoong, A.~Farhadi, Y.~Carmon, S.~Kornblith, and L.~Schmidt.
\newblock Model soups: averaging weights of multiple fine-tuned models improves
  accuracy without increasing inference time, 2022.
\newblock URL \url{https://arxiv.org/abs/2203.05482}.

\end{thebibliography}







\section{Appendix}
\vspace{-3mm}

\subsection{Relevancy as VLM's confidence}
\vspace{-2mm}

In our experiments, we have observed that directly using VLM relevancy maps works better than binarizing the relevancy maps with a cutoff threshold or clipping it.
The former collapse activation magnitudes and both are sensitive to the cutoff threshold.
In contrast, our design choice of using raw VLM relevancy maps contains the full range of relevancy activations.
We hypothesize that relevancy maps activation magnitudes give our models information about the VLM's confidence.

This interpretation informed our design choice of \ours's applications.
For instance, for Visually-Obscured Object Localization, the SemAbs module takes as input both the target and reference object relevancy point clouds, even when the target object is visually-obscured or hidden.
This allows the VLM to inform our networks when it thinks an object is not present in the scene (please refer to the \href{https://semantic-abstraction.cs.columbia.edu/}{project website} for examples).

\subsection{Network}
\vspace{-2mm}


We use a 3D U-Net~\cite{abdulkadir2016unet3d} architecture as $\threeDNet$, and a 2 layer MLP as $\ImplicitDecoder$.
For our scattering operation, we use max reduction, such that if multiple points are scattered into the same voxel, the voxel assumes the max of the points' features.
Our voxel grid has lower and upper bounds $(-1.0\si{\metre},-1.0\si{\metre},-0.1\si{\metre})$ and $(1.0\si{\metre},1.0\si{\metre},1.9\si{\metre})$ respectively.
We use random transformations (translation, rotation, and scale) on input and output point clouds, then filter points outside of the voxel grid bounds.



\subsection{Relevancy Extractor details}
\vspace{-2mm}

We batch parallelize along all crops (within each scale), scales, augmentations, and prompts.
Tuning the sliding window step size requires trading off running time with relevancy map quality.
In our experiments, we use step sizes a quarter of the crop size for each scale, which qualitatively gave decent relevancy maps while running in a reasonable amount of time.

In our experiments, we use multi-scale relevancy with 5 random RGB augmentations, horizontal flipping, crop sizes in the range $\{h,h/2,h/3,h/4\}$, where $h = 896$ is the image width, and strides one-fourth of their respective kernel sizes.
Our implementation takes $39.5\pm 0.1$s second for 100 labels (0.4 seconds per label) on this configuration.
In contrast, directly using \citeauthor{chefer2021generic}'s implementation in a sliding window fashion takes a total of $2420.8\pm 14.1$ seconds (19.3 seconds per text label)

We have released our multi-scale relevancy extractor on \href{https://github.com/columbia-ai-robotics/semantic-abstraction#multi-scale-relevancy-extractor}{Github} and hosted a (CPU-only) \href{https://huggingface.co/spaces/huy-ha/semabs-relevancy}{Hugging Face Spaces} for demo purposes.

\begin{figure}[t]
	\centering
	\includegraphics[width=\linewidth]{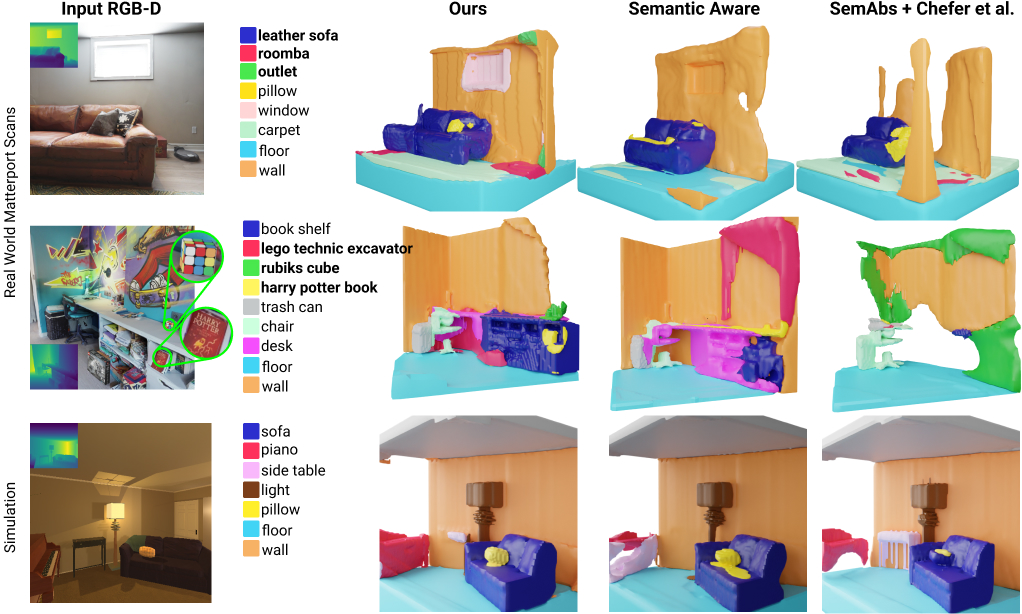}
	\caption{ \footnotesize\textbf{SSC Qualitative Comparisons.}}

	\label{fig:ssc_compare}
\end{figure}

\begin{figure}[t]
	\centering
	\includegraphics[width=\linewidth]{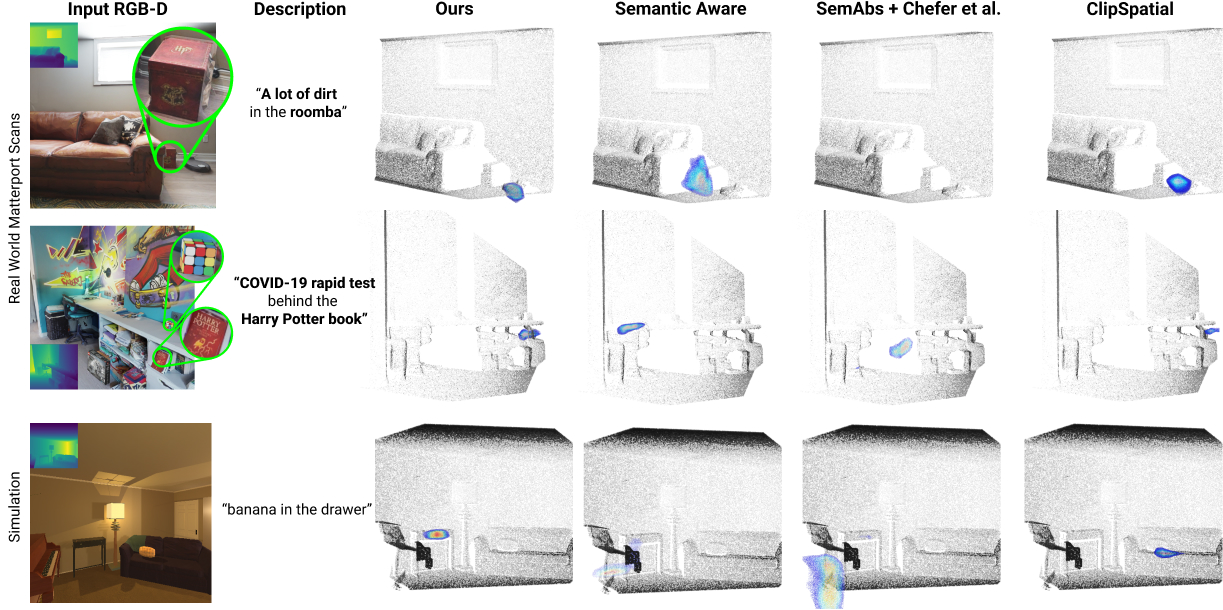}
	\caption{ \footnotesize
		\textbf{VOOL Qualitative Comparisons.}
	} \vspace{-5mm}

	\label{fig:vool_compare}
\end{figure}

\begin{figure}[t]
	\centering
	\includegraphics[width=\linewidth]{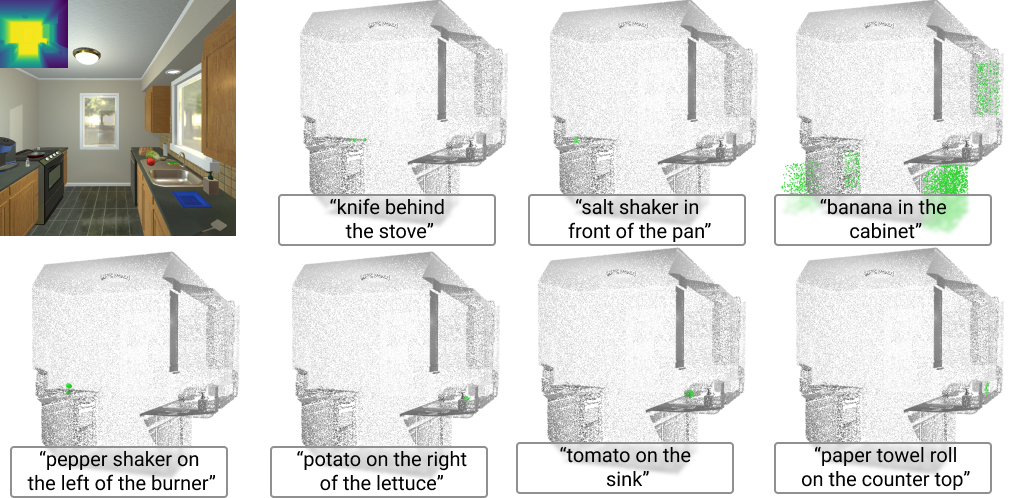}
	\caption{ \footnotesize
		\textbf{Ground truth VOOL labels.}
	} \vspace{-5mm}

	\label{fig:groundtruth_vool}
\end{figure}

\begin{figure}[t]
	\centering
	\includegraphics[width=\linewidth]{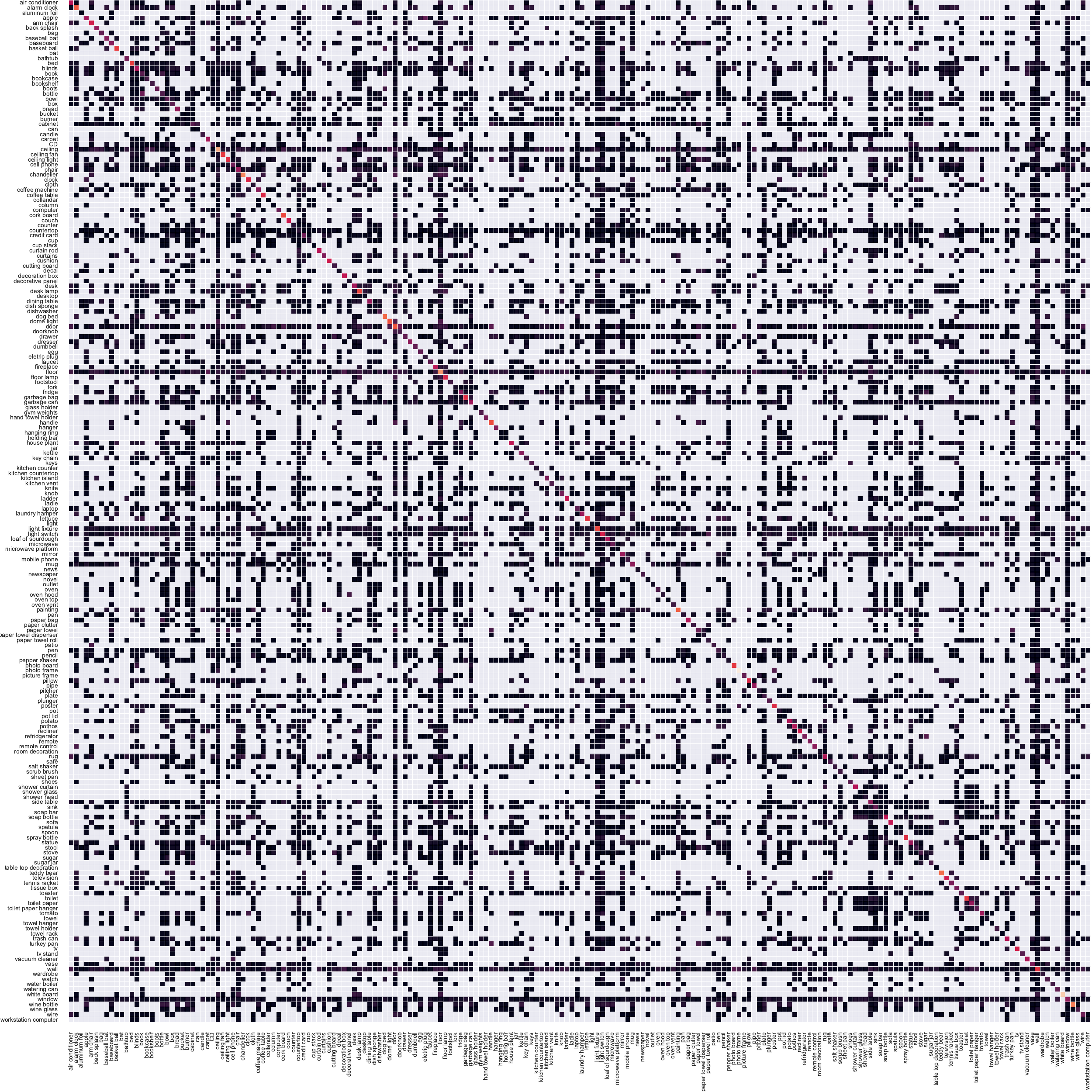}
	\caption{ \footnotesize
		\textbf{SemAbs's OVSSC Confusion Matrix.}
		Some of the biggest non-diagonal values include the pairs (door, door knob),(floor, carpet), (photo board, poster) and (photo frame, poster).
		Objects which don't co-occur in the same view in our testing dataset are colored white (\eg wine glass and toilet paper, white board and egg).
	} \vspace{-5mm}

	\label{fig:confusion}
\end{figure}



\subsection{Data Generation and Training Details}
\vspace{-3mm}


\textbf{OVSSC dataset.}
The training dataset contained 5063 views split across 100 scenes.
The evaluation dataset for novel rooms, novel visual, novel synonym, and novel classes contained 999, 999, 751, 1864 views respectively, split across the 20 test scenes.
We generate training data for $\threeDNet$ and $\ImplicitDecoder$ using our custom AI2-THOR~\cite{ai2thor} simulator.
Since the original simulator does not provide functionality to output 3D occupancies, we implement this with spherical collision detection for each query point.
To generate views in the rooms, we spawn the robot at random locations and Z-rotations and render RGB-D images, filtering views with too few objects.
In each batch, we sample $B$ scenes, $K$ classes within each scene, and $N$
points within each $\relevancypcd$. Using $M$ query points, SemAbs module's
occupancy prediction for each point is supervised to the ground-truth
occupancies using binary cross entropy (BCE), optimized using the AdamW
optimizer with learning rate 5e-4, a cosine annealing with warm restarts
learning rate scheduler. In our experiments, we use a $B = 4$, $K = 4$, $N =
	80000$, and $M = 400000$. Our 200 epoch training takes 2 days on a 4 NVIDIA
A6000's.

\textbf{VOOL dataset.}
The training dataset contained 6085 views split across 100 scenes.
The evaluation dataset for novel rooms, novel visual, novel synonym, and novel classes contained 1244, 1244, 940, 597 views respectively, split across the 20 test scenes.
We focus on six common spatial prepositions: behind, left of, right of, in front, on top of, and inside (Fig.~\ref{fig:groundtruth_vool}).
As in OVSSC (\S \ref{sec:method:ovssc}), we use our custom AI2-THOR~\cite{ai2thor} simulator to generate training data.
Specifically,  we define ``on top of'' and ``inside'' using AI2-THOR's receptacle information, while ``behind'', ``left of'', ``right of'', and ``in front'' are defined in a viewer-centric fashion.
Specifically, for these viewer-centric spatial relations, we first compute displacement between all pairs of objects (using their ground truth 3D occupancies).
Using these pairwise displacements, we determine if there is a spatial relation (\eg ``left of'') between each pair if their displacement is aligned with the direction (\eg dot product with the view-centric left direction) and if the pair's distance is small enough with respect to each pair's object dimensions.
The latter condition handles the intuition that spatial relations aren't usually defined in an absolute frame, but instead relative to the relevant objects (\eg a pen 1 meter to the right of an eraser is not "on the right of"  the eraser, but a tree 1 meter to the right of a house is "on the right of" the house).
To handle ambiguous descriptions (\eg ``banana in the cabinet'' when there are multiple cabinets), our ground truth positives contain all points consistent with the description (\eg points in the ``inside'' receptacle for all cabinets in view are labeled positive, Fig.~\ref{fig:groundtruth_vool}, top right).
We use the same hyperparameters and training setup as in OVSSC with the following exceptions. First, $K=6$ is the
number of descriptions per scene. Second, the scaled-cosine similarlity between
	{\small
		$\phi_Q^{{\threeDFeatureVolume_\textrm{target}\|\threeDFeatureVolume_\textrm{ref}}}$}
and $\spatialEmbeddingfunction(\spatialrelation)$ is supervised using BCE.

\textbf{Novel Semantic Class.}
We chose 6 test classes, which were held out during training for Novel Class evaluation.
To ensure we covered the ``inside'' spatial relation for novel classes, we included ``mug'' (small container), ``pot'' (medium container), and ``safe'' (larger container).
We also included ``wine bottle'',  ``teddy bear'' and ``basket ball''.
We chose these classes by looking at their naturally occurring frequency in views generated in Thor~\cite{ai2thor}, and chose classes that neither occurred too frequently (such that too many views will be held out for testing) nor too infrequently (such that all views of the class are in one or two Thor rooms).

\textbf{Novel Synonyms.}
For the following words, we replaced all of their occurences in text inputs (semantic class inputs for OVSSC, description for VOOL), with the following words which had a similar semantic meaning.
\vspace{-3mm}
\begin{multicols}{3}
	\begin{itemize}\setlength\itemsep{0.05em}
		\vspace{-5mm}
		\footnotesize
		\item television: tv
		\item sofa: couch
		\item house plant: plant in a pot
		\item bookcase: bookshelf
		\item baseball bat: rawlings big stick maple bat
		\item pillow: cushion
		\item arm chair: recliner
		\item bread: loaf of sourdough
		\item cell phone: mobile phone
		\item desktop: computer
		\item dresser: wardrobe
		\item dumbbell: gym weights
		\item fridge: refridgerator
		\item garbage can: trash can
		\item laptop: computer
		\item outlet: eletric plug
		\item stairs: staircase
	\end{itemize}
\end{multicols}
\vspace{-5mm}

\subsection{Things which did not work}

\mypara{Segmentation from Relevancy.}
Thresholding the raw relevancy activation maps would give a binary mask that can be interpreted as CLIP's segmentation for some class.
However, we hypothesize this performs poorly for three reasons.
First, relevancy activations have different magnitudes for different classes (\eg much stronger for ``plant'' than for ``wall'') and different views (\eg much stronger for a side view of a ``drill'' than a top down view of a ``drill'') which means a single threshold value doesn't work for all cases.
Second, relevancy highlights what a perception model ``looked'' at to make a certain prediction, which is rarely the entire object.
For instance, we observed that relevancy maps for ``table'' typically only highlight parts of the legs and not the entire object.
Lastly, relevancy activations also give information on the VLM's uncertainty.
While these raw values aren't interpretable, a neural network can be trained to extract information from these raw relevancy activation values.
This means that while raw activations aren't that useful for interpretability purposes, they can be used as input to a network just fine.

\mypara{Scaling Laws.}
The results for CLIP~\cite{radford2021learning} demonstrate that larger VITs demonstrate better zero-shot robustness.
We were hoping to show that using larger CLIP VIT models with the same training setup also exhibit the same performance scaling.
To our surprise, relevancy maps from any CLIP model other than the B/32 model didn't look promising.
This is a \href{https://github.com/hila-chefer/Transformer-MM-Explainability/issues/8}{known phenomenon} with the relevancy extraction approach we built upon.

\subsection{More results}
\vspace{-3mm}


\textbf{OVSSC Qualitative Comparison.}
In our qualitative OVSSC comparisons (Fig.~\ref{fig:ssc_compare}),  we observed that both baselines tend to perform poorly on small objects, such as ``outlet'' (first row) and ``rubiks cube''  (second row).
In addition, while the SemAware baseline can give reasonable predictions on training classes, such as floor, wall, and sofa, it struggles with novel classes, like ``roomba'' (completely absent, first row) or ``lego technic excavator'' (wrong prediction, second row).

\textbf{VOOL Qualitative Comparison.}
We show qualitative VOOL comparisons in Fig.~\ref{fig:vool_compare}.
The SemAware baseline struggled with descriptions containing unknown semantic classes (first two rows) and incorrectly identified the piano as a drawer (third row).
Given a suboptimal relevancy map as input, the SemAbs + \cite{chefer2021generic} baseline misses the small reference objects in all three cases and  predicted incorrect regions as a result.
Even ClipSpatial, our quantitatively strongest baseline, did not have enough information to properly learn spatial relations (second row, incorrectly predicted a region \emph{in front} of the book, not \emph{behind}) when spatial relations were also abstracted into relevancy maps.

\subsection{VOOL Performance Breakdown by Spatial Relation}
\vspace{-3mm}
We include a table of VOOL results, with performance divided up by each spatial relation in Table~\ref{tab:vool_full}.
We observed that our approach consistently outperforms all other approaches.

\begin{table}[t]
	\setlength{\tabcolsep}{0.055cm}
	\scriptsize
	\centering
	\begin{tabular}{l|l|rrrr}
		Approach                & Spatial Relation & Novel Room & Novel Visual & Novel Vocab & Novel Class \\
		\toprule
		Semantic Aware          & in               & 15.0       & 14.7         & 7.6         & 1.8         \\
		                        & on               & 9.0        & 8.9          & 11.4        & 4.5         \\
		                        & on the left of   & 11.2       & 11.1         & 14.4        & 4.0         \\
		                        & behind           & 12.8       & 12.6         & 14.1        & 2.2         \\
		                        & on the right of  & 13.1       & 13.0         & 11.5        & 3.4         \\
		                        & in front of      & 11.2       & 11.1         & 9.3         & 2.2         \\
		                        & mean             & 12.1       & 11.9         & 11.4        & 3.0         \\
		\midrule
		ClipSpatial             & in               & 9.6        & 8.6          & 7.1         & 3.3         \\
		                        & on               & 14.1       & 12.1         & 18.5        & 20.0        \\
		                        & on the left of   & 11.0       & 9.4          & 14.2        & 13.2        \\
		                        & behind           & 11.3       & 9.9          & 14.1        & 8.9         \\
		                        & on the right of  & 12.1       & 10.6         & 16.2        & 11.5        \\
		                        & in front of      & 12.3       & 10.3         & 15.7        & 9.9         \\
		                        & mean             & 11.7       & 10.1         & 14.3        & 11.2        \\
		\midrule
		SemAbs + [Chefer et al] & in               & 11.8       & 11.1         & 5.7         & 2.1         \\
		                        & on               & 7.0        & 6.7          & 11.3        & 7.1         \\
		                        & on the left of   & 9.5        & 9.3          & 13.7        & 4.9         \\
		                        & behind           & 7.6        & 7.6          & 10.6        & 2.5         \\
		                        & on the right of  & 9.2        & 9.2          & 11.0        & 3.9         \\
		                        & in front of      & 9.4        & 9.0          & 12.0        & 3.3         \\
		                        & mean             & 9.1        & 8.8          & 10.7        & 4.0         \\
		\midrule
		Ours                    & in               & 17.8       & 17.5         & 8.5         & 7.3         \\
		                        & on               & 21.0       & 18.0         & 27.2        & 28.1        \\
		                        & on the left of   & 22.0       & 20.3         & 27.7        & 25.1        \\
		                        & behind           & 19.9       & 18.0         & 22.8        & 16.7        \\
		                        & on the right of  & 23.2       & 21.7         & 28.1        & 22.1        \\
		                        & in front of      & 21.5       & 19.4         & 25.8        & 19.1        \\
		                        & mean             & 20.9       & 19.2         & 23.4        & 19.7        \\
	\end{tabular}
	\vspace{1mm}
	\caption{
		\textbf{Visually Obscured Object Localization by Spatial Relation.}
	}
	\label{tab:vool_full}
\end{table}



\end{document}